\documentclass[sigconf]{acmart}
\AtBeginDocument{%
  \providecommand\BibTeX{{%
    \normalfont B\kern-0.5em{\scshape i\kern-0.25em b}\kern-0.8em\TeX}}}

\copyrightyear{2023}
\acmYear{2023}
\setcopyright{acmcopyright}\acmConference[WSDM '23]{Proceedings of the Sixteenth ACM International Conference on Web Search and Data Mining}{February 27-March 3, 2023}{Singapore, Singapore}
\acmBooktitle{Proceedings of the Sixteenth ACM International Conference on Web Search and Data Mining (WSDM '23), February 27-March 3, 2023, Singapore, Singapore}
\acmPrice{15.00}
\acmDOI{10.1145/3539597.3570377}
\acmISBN{978-1-4503-9407-9/23/02}

\usepackage{xcolor}
\usepackage{bbm}
\usepackage{multicol}
\usepackage{multirow}
\usepackage{booktabs}
\usepackage{graphicx}
\usepackage{caption}
\usepackage{subcaption}
\usepackage{enumitem}
\usepackage[normalem]{ulem}
\useunder{\uline}{\ul}{}
\usepackage{makecell}
\usepackage{algorithmic}
\usepackage{amsmath}
\usepackage[ruled,linesnumbered,noend]{algorithm2e}

\DeclareUnicodeCharacter{2212}{-}
\begin{document}

\settopmatter{printacmref=true,printccs=true}

\title{Alleviating Structural Distribution Shift in \\
Graph Anomaly Detection}

\author{Yuan Gao}
\email{yuanga@mail.ustc.edu.cn}
\affiliation{%
  \institution{University of Science and Technology of China}
  \country{}}
  
\author{Xiang Wang}
\authornote{Corresponding authors}
\email{xiangwang1223@gmail.com}
\affiliation{%
  \institution{University of Science and Technology of China}
  \country{}}
  
\author{Xiangnan He}
\authornotemark[1]
\email{xiangnanhe@gmail.com}
\affiliation{%
  \institution{University of Science and Technology of China}
  \country{}}
  
\author{Zhenguang Liu}
\email{liuzhenguang2008@gmail.com}
\affiliation{%
  \institution{Zhejiang University}
  \country{}}
  
\author{Huamin Feng}
\email{oliver_feng@yeah.net}
\affiliation{%
  \institution{Beijing Electronic Science and Technology Institute}
  \country{}}
  
\author{Yongdong Zhang}
\email{zhyd73@ustc.edu.cn}
\affiliation{%
  \institution{University of Science and Technology of China}
  \country{}}

\renewcommand{\shortauthors}{Yuan Gao et al.}

\newcommand{\ie}{\emph{i.e., }}
\newcommand{\eg}{\emph{e.g., }}
\newcommand{\ifff}{\textit{iff. }}
\newcommand{\etal}{\emph{et al.}}
\newcommand{\st}{\emph{s.t. }}
\newcommand{\etc}{\emph{etc.}}
\newcommand{\wrt}{\emph{w.r.t. }}
\newcommand{\cf}{\emph{cf. }}
\newcommand{\aka}{\emph{aka. }}

\newcommand\norm[1]{\left\lVert#1\right\rVert}
\newcommand{\Lapl}{\mathbf{\mathop{\mathcal{L}}}}
\newcommand{\lapl}{\mathcal{L}}
\newcommand{\Trans}[1]{{#1}^{\top}}
\newcommand{\Trace}[1]{tr\left({#1}\right)}
\newcommand{\Bracs}[1]{\left({#1}\right)}
\newcommand{\Mat}[1]{\mathbf{#1}}
\newcommand{\MatS}[3]{\mathbf{#1}^{#2}_{#3}}
\newcommand{\Space}[1]{\mathbb{#1}}
\newcommand{\Set}[1]{\mathcal{#1}}
\newcommand{\vectornorm}[1]{\left|\left|#1\right|\right|}
\newcommand{\std}[1]{\scriptsize{$\pm$#1}}
\newcommand{\boldit}[1]{\textbf{\textit{#1}}}

\begin{abstract}
    Graph anomaly detection (GAD) is a challenging binary classification problem due to its different structural distribution between anomalies and normal nodes --- abnormal nodes are a minority, therefore holding high heterophily and low homophily compared to normal nodes. Furthermore, due to various time factors and the annotation preferences of human experts, the heterophily and homophily can change across training and testing data, which is called structural distribution shift (SDS) in this paper. The mainstream methods are built on graph neural networks (GNNs), benefiting the classification of normals from aggregating homophilous neighbors, yet ignoring the SDS issue for anomalies and suffering from poor generalization.
     
    This work solves the problem from a feature view. We observe that the degree of SDS varies between anomalies and normal nodes. Hence to address the issue, the key lies in resisting high heterophily for anomalies meanwhile benefiting the learning of normals from homophily. Since different labels correspond to the difference of critical anomaly features which make great contributions to the GAD, we tease out the anomaly features on which we constrain to mitigate the effect of heterophilous neighbors and make them invariant. However, the prior distribution of anomaly features is dynamic and hard to estimate, we thus devise a prototype vector to infer and update this distribution during training. For normal nodes, we constrain the remaining features to preserve the connectivity of nodes and reinforce the influence of the homophilous neighborhood. We term our proposed framework as \textit{\textbf{G}raph \textbf{D}ecomposition \textbf{N}etwork} (GDN). Extensive experiments are conducted on two benchmark datasets, and the proposed framework achieves a remarkable performance boost in GAD, especially in an SDS environment where anomalies have largely different structural distribution across training and testing environments. Codes are open-sourced in \url{https://github.com/blacksingular/wsdm_GDN}.
\end{abstract}



\maketitle

\section{Introduction}
Anomalies (\aka fraudsters) delineate the abnormal objects that deviate significantly from the normal (\aka benigns) \cite{ma2021comprehensive}.
Detecting anomalies has attracted considerable attention in many real-world domains, such as identifying spams in reviews \cite{dou2020enhancing}, misinformation in social networks \cite{cheng2021causal,gao2022rumor}, and frauds in financial transactions \cite{liu2021pick}.
In general, abnormal and normal objects are interdependent with rich relationships, which can be naturally organized as graphs \cite{dou2020enhancing}.
Wherein, nodes represent these objects, and edges interpret their relationships.
On such structural data, leading methods \cite{dou2020enhancing,liu2021pick,zhangfraudre,chang2021f} frame the graph abnormal detection (GAD) problem as the semi-supervised node classification task, where only a fraction of nodes are labeled as the training data, and the remaining nodes form the testing data.
To distill the discriminative information for the hidden anomalies, these methods mostly apply graph neural networks (GNNs) \cite{Kipf2017,hamilton2018inductive,veli2018graph} that propagate the label-aware signals along with the graph structure.

\begin{figure}[!]
    \centering
    \includegraphics[width=\linewidth]{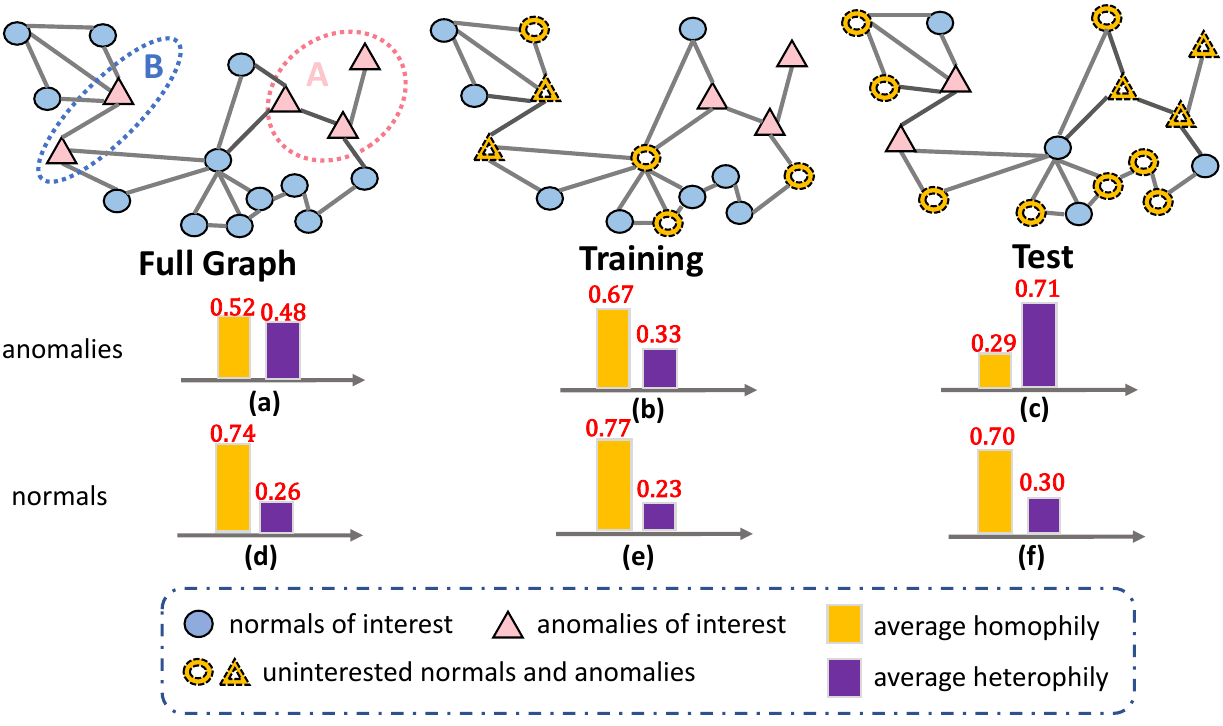}
    \caption{Illustration of SDS in GAD. Cells in the dotted line mean that they are of no interest in the current environment (training or test). The average homophily and heterophily for anomalies and normals are presented.}
    \label{fig:ood}
    \vspace{-15pt}
  \end{figure}

Here we take a closer look at GAD, especially the structural distribution \wrt heterophily and homophily.
Specifically, \textbf{heterophily} \cite{rogers1970homophily} indicates the phenomenon that edges connect the nodes from different classes (\ie the anomaly and normal classes), which contrasts with \textbf{homophily} that counts for edges between the same-class nodes.
Clearly, for a node, the distribution \wrt heterophily and homophily shows the label information within its local neighborhood, as shown in Figure \ref{fig:ood}(a).
The distribution difference between anomaly and normal nodes is amplified by GAD's class imbalance nature --- that is, anomalies are in the minority subgroups and submerged in the normal nodes, thus easily holding higher heterophily than normals, as the comparison between Figures \ref{fig:ood}(a) and (d) shows.
Hence, it is of importance to differentiate the structural distributions of anomaly and normal nodes, so as to refine the class-wise signals better.

Furthermore, we find that conducting GAD in a semi-supervised manner naturally faces the challenge of structural distribution shift (SDS) --- that is, the structural distribution \wrt heterophily and homophily can change across the training and testing datasets.
For example, as compared between Figures \ref{fig:ood}(b) and (c), heterophily of anomalies is $0.67$ and $0.29$ in the training and testing environments, respectively;
in contrast, heterophily of normal nodes only is much more stable across two different environments, as Figures \ref{fig:ood}(e) and (f) show.
Here we present several insights into SDS:
\begin{itemize}[leftmargin=*]
    \item SDS happens in the open environment, which usually includes different distributions that are collected with various time factors \cite{wu2022towards}, annotation preferences \cite{he2018instance}. Hence, robustness to SDS is critical for deploying GAD models in real-world scenarios.
    \item Distinguishing anomalies from the normal suffer from the shift \wrt heterophily. Specifically, as most GNN-based GAD models \cite{dou2020enhancing,liu2021pick,zhangfraudre} blindly aggregate the neighborhood information, the representations of testing anomalies absorb more information from normal neighbors than that of training anomalies, thereby inundating the crucial cues. Therefore, these GAD models will generalize poorly in the testing anomalies.
    \item In contrast to anomalies, normals hold more stable distributions across the training and testing sets. Thus, classifying normals can benefit from the class-aware patterns of homophilous neighbors.
\end{itemize}

However, alleviating SDS in GAD remains largely unexplored, but is the focus of our work. Specifically, most early studies \cite{veli2018graph,Kipf2017,hamilton2018inductive} blindly employ GNNs to perform information propagation among nodes, without inspecting the influence of different neighbors. Some follow-up works \cite{dou2020enhancing,liu2021pick,zhangfraudre} control the information being propagated by discarding some edges based on the neighbor similarity.
Although these studies help mitigate the heterophily gap between anomalies and normals, they leave the SDS issue untouched. Thus, they struggle to fit the testing anomalies with the learned patterns from the training nodes.

In this paper, we argue that the key to alleviating SDS is differentiating structural patterns for anomalies and normals. We assume that (partial) anomaly features are quite useful for GAD\cite{DBLP:journals/corr/abs-2202-07902}. These features have high variance across anomalies and normal nodes, therefore in which dimensions nodes are more likely to absorb noisy signals from heterophilous neighbors. This observation leads us to the feature disentanglement. Inspired by variable decomposition in stable learning \cite{shen2020stable, fan2022debiased}, we devise a new framework, \underline{G}raph \underline{D}ecomposition \underline{N}etwork (GDN). Specifically, for anomalies, it tries to identify the anomaly pattern which is made invariant to SDS, so as to reduce the negative influence of heterophily shift; meanwhile, for normals, it attempts to extract the pattern retaining to benefit of homophily.
To achieve these two strategies, it resorts to constraints on node features and divides them into two parts: class and surrounding features. Class features are constrained to approach the prior distribution of node features, so as to prevent anomalies from absorbing noisy signals from the neighborhood and influences from heterophily shift; meanwhile, surrounding features preserve the connectivity of two neighboring nodes, to enhance the information of homophilous neighbors. This allows us to identify anomaly features invariant to heterophily shift and capture the local homophily of normals, thus boosting the overall GAD performance.
\section{Preliminaries}
In this section, we illustrate the task of GAD and the imbalanced heterophily property.

\vspace{5pt}
\noindent \textbf{Graph Anomaly Detection.} Conventional anomaly detection techniques always consider isolated data instances while ignoring the relationship between instances which carries complementary information \cite{akoglu2015graph}. Taking spam review for e-commerce as an example, multiple relations can be established between reviews, \eg reviews posted by the same user, so as those posted under the same item. In this manner, we reorganize the anomalies and normals as an attributed multi-relation graph, which can be defined as $\mathcal{G}=\{\mathcal{V}, \{\mathcal{E}_{r}\}, \Mat{X}\}$. $\mathcal{V}$ denotes a set of anomaly and normal nodes; $\mathcal{E}_{r}$ stands for edges \wrt relation $r \in \{1,2,...,R\}$, which can be rules, interactions, or shared attributes between nodes \cite{dou2020enhancing}; $\Mat{X}$ is the attribute matrix, each row of which is a $d$-dimensional vector representing the features of the corresponding node. 

GAD has been addressed as a semi-supervised node classification task. Most of the time, anomalies are regarded as positive with label 1, while normal nodes are seen as negative with label 0. The whole graph contains two types of nodes, $\mathcal{V}_{train}$ and $\mathcal{V}_{test}$. $\mathcal{V}_{train}$ are labeled with $\Mat{Y}_{train}$, while the labels $\Mat{Y}_{test}$ are inaccessible during training. Formally, leading solutions \cite{dou2020enhancing, liu2021pick, zhangfraudre} employ GNN as the predictive model $f$ to achieve small error on predicting the ground truth $\Mat{Y}_{test}$ for unobserved nodes $\mathcal{V}_{test}$
\begin{equation}
    f(\mathcal{G}, \Mat{Y}_{train}) \rightarrow \hat{\Mat{Y}}_{test}.
\end{equation}

\vspace{5pt}
\noindent \textbf{Heterophily and homophily.} Given a set of nodes along with their labels, the heterophily of a node can be defined as the ratio of the edges connecting the nodes in different classes (\ie the anomaly and normal classes). For each node, the sum of its heterophily and homophily degree equals 1:
\begin{equation}
\begin{aligned}
    X_{hetero}(v) &= \frac{1}{|\mathcal{N}(v)|}|\{u: u \in \mathcal{N}(v), y_u \neq y_v\}|,\\
    X_{homo}(v) &= \frac{1}{|\mathcal{N}(v)|}|\{u: u \in \mathcal{N}(v), y_u = y_v\}|,
\end{aligned}
\end{equation}
\noindent where $\mathcal{N}(\cdot)$ stands for the neighborhood of a central node. In GAD, enhanced by the imbalance nature, anomalies have high heterophily, while normal nodes have relatively low heterophily. An example on real dataset is shown in Table \ref{tab:hetero}, from which we observe this phenomenon, as well as the shift \wrt heterophily and homophily across training and test environments.

\begin{table}[!]
\caption{Heterophily and homophily in Amazon}
\label{tab:hetero}
\begin{tabular}{c|cc|cc}
\hline
\multirow{2}{*}{} & \multicolumn{2}{c|}{Training}                   & \multicolumn{2}{c}{Test}                     \\ \cline{2-5} 
                  & \multicolumn{1}{c|}{Homo} & Hetero & \multicolumn{1}{c|}{Homo} & Hetero \\ \hline
Anomalies         & \multicolumn{1}{c|}{0.113}     & 0.887       & \multicolumn{1}{c|}{0.043}     & 0.957       \\ \hline
Normal            & \multicolumn{1}{c|}{0.979}     & 0.021       & \multicolumn{1}{c|}{0.976}     & 0.024       \\ \hline
\end{tabular}
\end{table}

\section{Methodology}
In this section, we present the details of the proposed framework GDN. We will first introduce the Structural Distribution Shift (SDS) problem in GAD. Then we elaborate on gradient-based invariant feature extraction as well as two constraints that guide the process.
\subsection{Structural Distribution Shift in GAD} We next formulate the structural distribution shift problem and figure out two major concerns: \textit{how will SDS affect the learning of GNNs and what causes the problem}?
\subsubsection{Definition}
SDS means the label distribution of the neighborhood is different between $p_{train}$ and $p_{test}$, which leads to different homophily and heterophily degrees for each node $v$:
\begin{equation}
\label{eq:label_shift}
p_{train}([X_{hetero}(v_{1}), X_{homo}(v_{1})]) \neq p_{test}([X_{hetero}(v_{2}), X_{homo}(v_{2})]).   
\end{equation}
It is usually caused by human annotation: anomalies with more intra-class edges are easier to be marked out. For easier expression, we denote the joint probability distribution $[X_{hetero}(v), X_{homo}(v)]$ as $\Psi(v)$, then the probability of an instance $v$ being labeled not only depends the node feature $x_{v}$ but on $\Psi(v)$:
\begin{equation}
\label{eq:shift}
p_{train}(\Psi(v),x_{v},o) \neq p(x_{v}, o), 
\end{equation}
where $p$ is the real distribution of samples, $o$ denotes observed training nodes. The extent of SDS can be defined as:
\begin{equation}
    \mathcal{D}[p_{train}(\Psi(v),x_{v},o), p(x_{v}, o)],
\end{equation}

\noindent where $\mathcal{D}(p(x),g(x))=\int_{\mathcal{X}}p(x)\psi(\frac{g(x)}{p(x)})dx$ is a distance function in Csiszár family, which is set as $KL$ divergence in this work.

\begin{figure}[!]
  \centering
  \includegraphics[width=\linewidth]{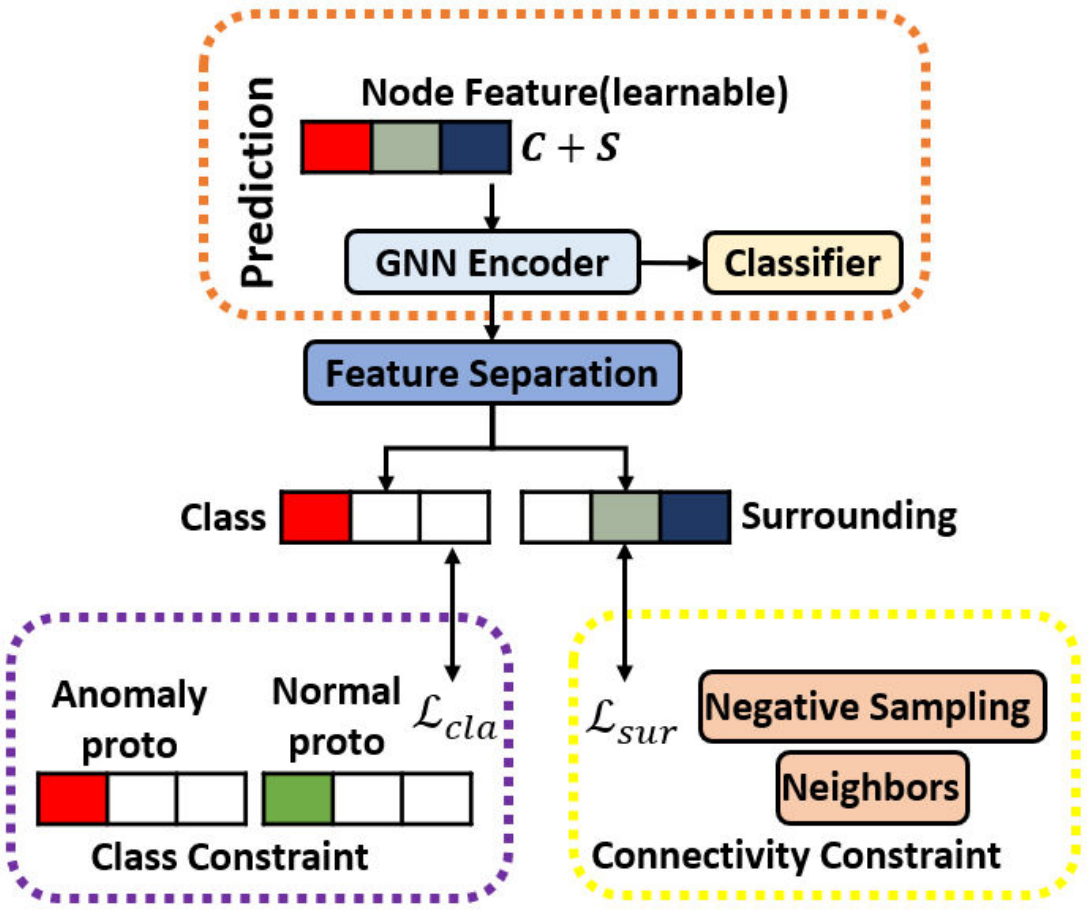}
  \caption{Illustration of GDN. The feature separation module separates the node feature into two sets. Two constraints are leveraged to assist separation. Blank positions in node representation mean they are zero when calculating losses.}
  \label{Figure3}
\vspace{-10pt}
\end{figure}

\subsubsection{Effects of Structural Distribution Shift on GAD} 
GNN-based GAD methods capture the neighborhood pattern and transfer it to unseen nodes, which ignore the SDS. This distribution inconsistency hinders the model from optimizing the ideal loss function which could definitely harm the model's performance \cite{zadrozny2004learning}:

\begin{equation}
    \Space{E}_{v\sim p}[l(g(\Psi(v),x_{v}))] \neq \Space{E}_{(v,o)\sim p_{train}}[l(g(\Psi(v),x_{v}))|o=1],
\end{equation}

\noindent where $\Space{E}{v\sim p}[l(g(\Psi(v),x_{v}))]$ is the expected value of the objective function for GNN-based methods over the distribution of all examples, and the right-hand side expectation is the expected value of loss solely on training (observed) set. For semi-supervised learning, the right-hand side expectation is directly minimized since only nodes with $o=1$ are available. In GAD, anomalies only occupy the minority of nodes and hold vast normal neighbors, making it a severe SDS environment. More specifically, anomalies, as the detection target, tend to have more anomaly neighbors in the training set, while connecting with more normal nodes in the testing. In an extreme case, all of the anomaly-anomaly edges are contained in the training set, and the neighbor label distribution of anomalies is totally skewed in the test set. Therefore, the ideal loss function can not be directly optimized, GNN classifier fails to capture the real underlying distribution of data samples.

\subsubsection{A closer look in SDS} \label{effects of SDS}
We present here an improved analysis through a closer look at the cause of SDS. First, we need to make an assumption about the probability distribution $P(O_i)$, which denotes whether the node is labeled, \ie a node is labeled if $O_i=1$, and $O_i=0$ otherwise. Following \cite{zadrozny2004learning}, we adopt a biased selection method to construct examples that have the SDS problem between training and testing environments. As discussed above, for node $i$, the probability of it being labeled is controlled by its neighborhood label similarity:
\begin{equation}
    P(O_i=1)=|\{j|j \in \mathcal{N}_i, y_i = y_j\}|/|\mathcal{N}_i|.
    \label{eq:bias_split}
\end{equation}
In this manner, if the homophily between central node $i$ with its neighborhood is larger, node $i$ will have a larger probability to appear in the training set. This setting is consistent with our cognition of the real world: an anomaly surrounded by more anomalies is easier to be identified.

Consider one-hop neighbor set $\mathcal{N}(v_{i})$ of a given central node $v_{i}$, then for binary classification problem, label distribution of each node $v_{j}$ in $\mathcal{N}(v)$ obeys $\Mat{y}\sim Bern(p_{j})$,  where $p_{j}$ is the probability of $v_{i}$ and $v_{j}$ have the same label. Then $\Psi(v)$, \ie the label distribution of $\mathcal{N}(v)$, is a family of probability distributions whose domain is bounded between 0 and 1, which leads us to Proposition 1.

\begin{figure}
  \begin{subfigure}[t]{.45\linewidth}
    \centering
    \includegraphics[width=\linewidth]{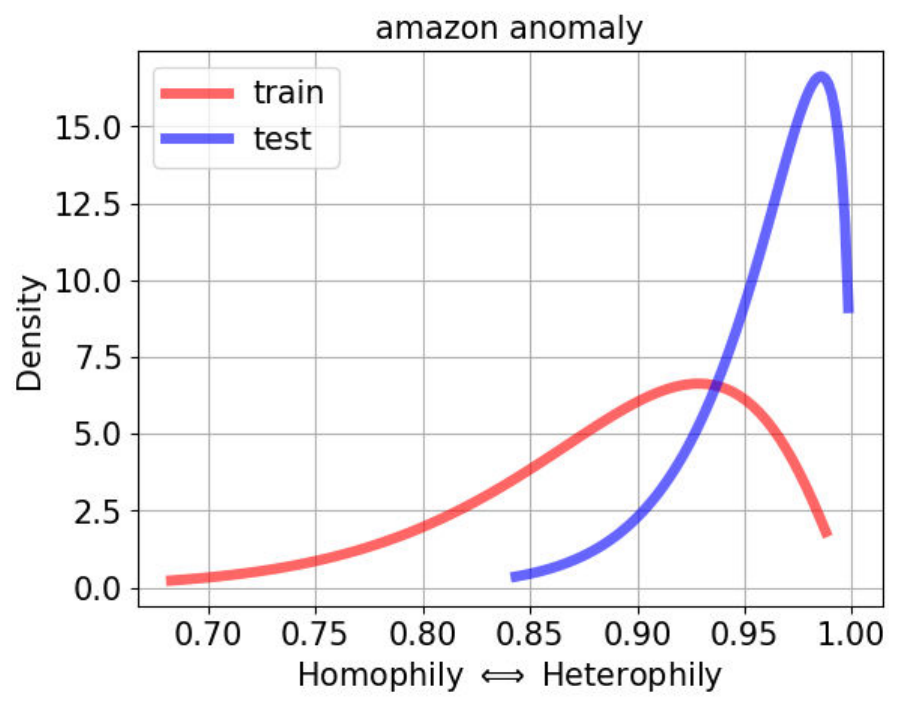}
    \caption{\textbf{Anomalies in Amazon}}
    \label{ano_ama}
  \end{subfigure}
  \hfill
  \begin{subfigure}[t]{.45\linewidth}
    \centering
    \includegraphics[width=\linewidth]{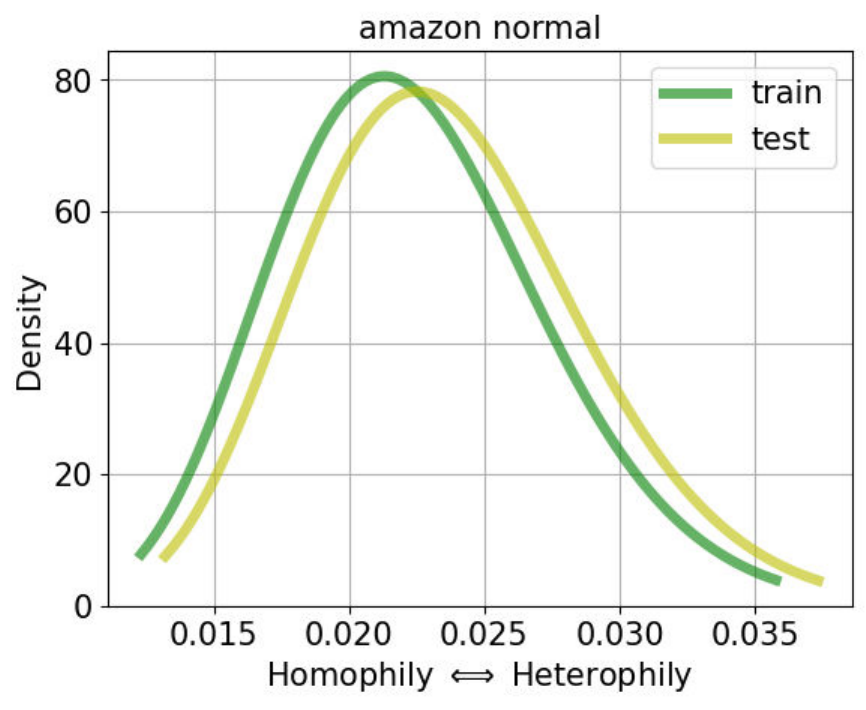}
    \caption{\textbf{Normals in Amazon}}
    \label{norm_ama}
  \end{subfigure}

  \medskip

  \begin{subfigure}[t]{.45\linewidth}
    \centering
    \includegraphics[width=\linewidth]{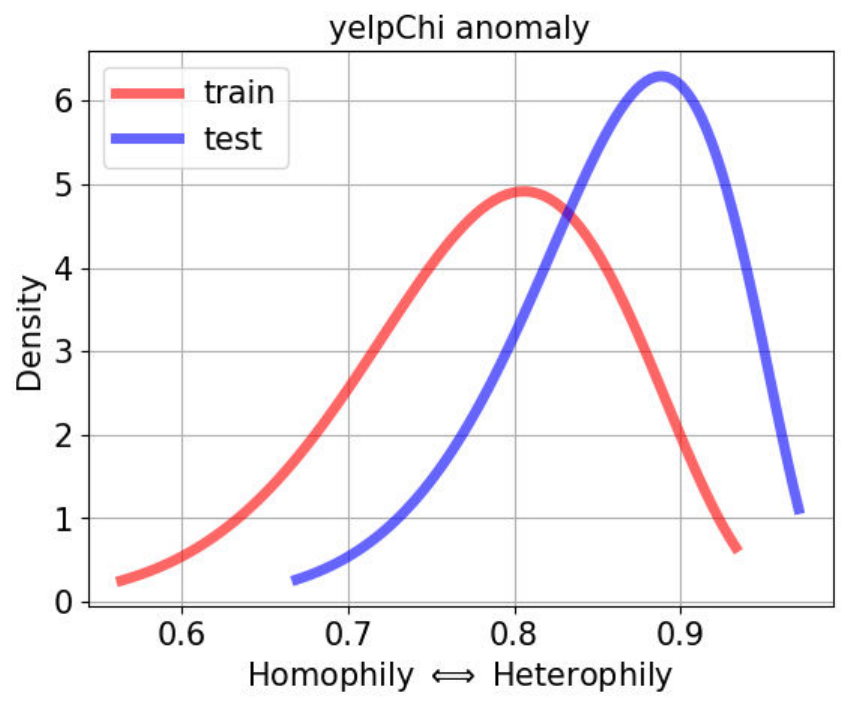}
    \caption{\textbf{Anomalies in YelpChi}}
    \label{ano_yelp}
  \end{subfigure}
  \hfill
  \begin{subfigure}[t]{.45\linewidth}
    \centering
    \includegraphics[width=\linewidth]{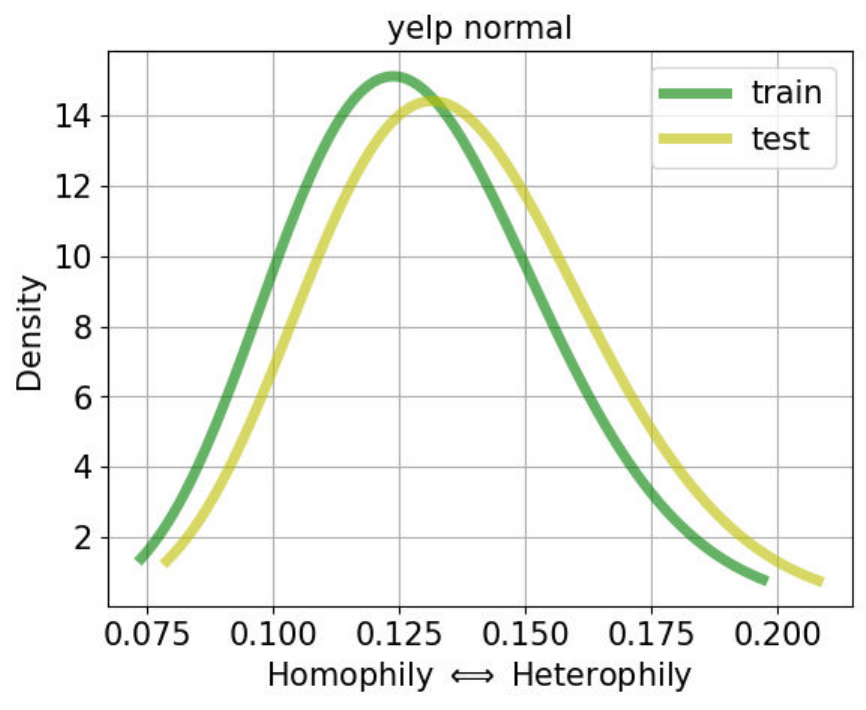}
    \caption{\textbf{Normals in YelpChi}}
    \label{norm_yelp}
  \end{subfigure}
  \caption{Visualization of \textbf{SDS} for different classes of nodes on two real world datasets. On x-axis, left means higher homophily and right means higher heterophily.}
  \label{visual_SDS}
  \vspace{-10pt}
\end{figure}

\vspace{5pt}
\noindent \textbf{Proposition 1} The label neighborhood distribution of a given node should obey $\Mat{Y}\sim Be(\alpha, \beta)$.
\vspace{5pt}

\noindent where $Be(\alpha, \beta)$ is the beta distribution parameterized by shape parameters $\alpha$ and $\beta$. With the above proposition, we split two real-world spam-review detection datasets according to this assumption based on which we conduct further analysis. The visualizations of $\Psi(v)$ on training and test are presented in Figure \ref{visual_SDS}. From the figure, we have two observations. (1) Comparing \ref{ano_ama} and \ref{norm_ama} (\ref{ano_yelp} and \ref{norm_yelp}), the structural distribution shift on anomalies is apparent, while that on normals is trivial, which indicates that different learning strategies should be adopted for different class nodes. (2) The extent of SDS is positively correlated with the extent of heterophily presented in Table \ref{tab:hetero}, which suggests that heterophily may be the cause of SDS.

\subsection{Invariant Feature Extraction}

In Section \ref{effects of SDS}, we inspect the cause of SDS, and suppose that imbalanced heterophily would exacerbate the problem. To alleviate the negative effect of SDS, the key is to identify a pattern expected to be invariant to SDS for anomalies. Intuitively, we want to reduce the neighbor label influence for anomalies. However, in GNNs the information is directly propagated through node features instead of labels, hence we want to bridge the gap between label influence and feature influence. Luckily, the recent method \cite{wang2021combining} theoretically formulates the relationship between them, and shows the edge weights which serve to aggregate node features also aggregate node labels over its immediate neighbors. From this perspective, we can mitigate the neighbor label influence for anomalies by reducing feature influence. Furthermore, we assume that (partial) features are useful for detection \cite{DBLP:journals/corr/abs-2202-07902}, \ie the distinguishment of anomalies. These features are quite different across different class nodes, therefore they are more likely to absorb noisy signals from heterophilous neighbors. This observation leads us to the invariant feature extraction.

For anomalies, inspired by variable decomposition \cite{shen2020stable, fan2022debiased}, we assume that node feature $X$ can be decomposed into \textit{class feature} $C$ and \textit{surrounding feature} $S$. We hope $C$ inherits most of the node informative characteristics which depict ``what the anomaly prototype looks like regardless of heterophily'', this set of features will be constrained by prototype (introduced in section \ref{sec:constraint}) which prevent the representation from deviating to heterophily neighbor signals. Formally, for variable $X_{train}, Y_{train}$ on training distribution $p_{train}$:
\begin{equation}
\begin{aligned}
\vspace{-2pt}
    \Space{E}[Y_{train}|\Phi(X_{train})] &= \Space{E}[Y_{train}|C_{train}, S_{train}] \\ &=\Space{E}[Y_{train}|C_{train}]=\Space{E}[Y|C],
\vspace{-2pt}
\end{aligned}
\end{equation}

\noindent where $\Phi(X_{train})$ is the neighbor feature distribution of training nodes. Note that in GNNs, a node itself is always included in $\Phi(X_{train})$ by adding self-loop, as $C$ inherits most of the node informative characteristics, $S$ is independent to label $Y$ given $C$. According to the assumption, $C$ is a property of node and is invariant to neighbors, it is unlikely affected by heterophily and SDS, \ie $\Space{E}[Y_{train}|C_{train}]=\Space{E}[Y|C]$. 

For normals, since they have low heterophily and hence their neighborhood information is more constant and cleaner, we expect that they aggregate neighborhood information which assist their own representation learning. We hope $S$ can capture local structure near normal nodes and summarize ``how normal nodes benefit from the low heterophily'', a connectivity constraint(introduced in section \ref{sec:constraint}) is applied to ensure neighbors share similar $S$. We next introduce how to appropriately separate features to meet our demand above.

Recall that we aim to extract the most informative $C$, intuitively, as the spirit of gradient descent, neural importance can be quantified as its absolute score of gradient value \wrt prediction loss. Following recent works \cite{chen2020counterfactual,pope2019explainability}, which utilize Grad-CAM \cite{selvaraju2017grad} or its variants to obtain local contribution to classification, we formulate the contribution of the $k$-th feature to the anomaly detection at layer $l$ as:
\vspace{-6pt}
\begin{equation}
    \alpha_{k}^{(l,c)} = \frac{1}{N}|\sum_{n=1}^N\frac{\partial y^{(c)}}{\partial H_{k,n}^{(l)}}|,
    \label{eq:gfs}
\vspace{-4pt}
\end{equation}

\noindent where $y^{(c)}$ is the predicted probability of ground truth $c$, H is the hidden layer feature representations and N is the total number of samples. We design a feature selector on the basis of this gradient score, which adaptively teases out class feature $C$ using top-K sampling, and leaves the rest as surrounding feature $S$. During the selection process, to ensure $C$ and $S$ meet expectations, we next introduce two constraints on the learning objectives.

\subsection{Constraints} \label{sec:constraint}
 Towards the goal of invariant feature extractions, we revise the learning objective of GAD by enforcing:  1) \textit{class constraints}: maximize the similarity of $C$ between two nodes if they belong to the same class, while minimizing it when they have the different labels, and 2) \textit{connectivity constraints}: maximize the similarity of $S$ between two nodes if they are neighbors, otherwise minimize it. We term these two constraints as $\mathcal{L}_{cla}$ and $\mathcal{L}_{sur}$, and introduce them as follows:


\begin{equation}
\begin{aligned}
    \mathcal{L}_{cla}(v) &= KL(C_{v}, proto_{+}) - KL(C_{v}, proto_{-}), \\
    \mathcal{L}_{sur}(v) &= \sum_{u\in \mathcal{N}(v)}KL(S_{u}, S_{v}) - \sum_{u\notin \mathcal{N}(v)}KL(S_{u}, S_{v}), \\
    \mathcal{L}_{constraint} &= \frac{1}{|N|}\sum_{v:y_{v}=1}\mathcal{L}_{cla}(v) + \mathcal{L}_{sur}(v),
\end{aligned}
\label{eq:cons}
\end{equation}

\noindent where $proto_{+}$ and $proto_{-}$ are prior distributions for anomalies and normals respectively, and the distributions depict "what the prototype looks like" for each class; $KL$ is the KL-divergence between two distributions. $\mathcal{L}_{sur}$ is easy to handle, by randomly sampling non-neighbors from the dataset and computing the distances between neighbors and non-neighbors respectively. Now we have transformed the problem into distribution estimation. A prevalent way is to learn a class-specific global context that gives a broad overview of the given class, namely the prototype vector. In our model, we acquire this prototype vector $proto$ adaptively. In every epoch, we register the current prototype $proto^{(e)}$, based on which we calculate the similarity between each node and $proto$. Nodes that deviated from $proto$ from this epoch should have a lower weight in the next update step \cite{chen2021gccad}. Formally, for nodes in each class:
\vspace{-5pt}
\begin{equation}
\label{eq:de}
\begin{aligned}
    s_v^{(e)} = cosine(h_v^{(e)}, proto^{(e-1)}), \quad &w_v^{(e)} = \frac{exp(s_v^{(e)}/\tau)}{\sum_{u=1}^{N}exp(s_u^{(e)}/\tau)} \\
    proto^{(e)} = &\sum_{v=1}^{N}w_{v}\cdot h_v^{(e)},
\end{aligned}
\vspace{-5pt}
\end{equation}
where $\tau$ is the temperature parameter to control the smoothness of weights. First, we compute the cosine similarity between each node $v$ and previous prototype vector $proto^{(e-1)}$, whose softmax output is served as weight $w_{v}^{(e)}$ for each node. Then the prototype $proto^{(e)}$ is updated by aggregating different nodes accordingly on the basis of this contribution. In practice, we initialize $proto^{(0)}$ as the average pooling of the vectors in the given class. The prototype vector serves as a constraint to endow class feature $C$ with resistance to heterophily and SDS.

\subsection{Final Loss Function}
Figure \ref{Figure3} presents an overview of our proposed model GDN. Following leading solutions, we adopt RGCN \cite{schlichtkrull2018modeling} as our backbone. For each node $v$, we define its final embedding as the output of the RGCN at the last layer $z_{v} = h_{v}^{(l)}$. And we leverage our constraint on cross-entropy as the final loss function for optimization:
\begin{equation}
\label{eq:lgnn}
    \mathcal{L}_{GDN} = \sum_{v \in \mathcal{V}}-log(y_v \cdot \sigma(z_v)) + \lambda exp(\mathcal{L}_{constraint}),
\end{equation}

where $\lambda$ is a hyper-parameter to control the balance between two losses. Note that node vectors should locate closer to same-class prototype than other-class one, thus $\mathcal{L}_{cla}$ is always negative; likewise, $S$ of neighbor nodes should look more similar than those of non-neighbor nodes, therefore, $\mathcal{L}_{sur}$ is less than zero; then the sum of two terms $\mathcal{L}_{constraint}$ is negative, which we transform to the exponential space for the sake of positivity of total loss.

\section{Experiments}
In this section, we conduct experiments on real-world datasets and report the results of our models as well as some state-of-the-art baselines to show the effectiveness of our proposed model. Particularly, we mainly aim to answer the following research questions:
\begin{itemize}[leftmargin=*]
    \item \textbf{RQ1:} How does our model GDN perform compared with SOTA graph anomaly detection methods?
    \item \textbf{RQ2:} Can GDN alleviate the SDS problem on biased dataset?
    \item \textbf{RQ3:} Is the proposed model effective under different hyper-parameter settings? 
    \item \textbf{RQ4:} Is GDN framework flexible to various GNN backbones and helpful in enhancing them? 
\end{itemize}

\subsection{Experimental Setup}

\subsubsection{Dataset}
Following previous works, We use the YelpChi dataset \cite{rayana2015collective} and Amazon dataset \cite{mcauley2013amateurs} to study GNN-based
fraud detection problem. The YelpChi dataset includes hotel and restaurant reviews filtered and recommended by Yelp. The Amazon dataset includes product reviews under the Musical Instruments category. Similar to \cite{dou2020enhancing}, we label users with more than 80\% helpful votes as benign entities and users with less than 20\% helpful votes as fraudulent entities. Both of the datasets are attributed multi-relation graph. Nodes in YelpChi dataset have 32-dimension features and 3 relations, and nodes in Amazon have 25-dimension features and 3 relations. For exact definition of these relations, we refer the readers to \cite{dou2020enhancing,liu2021pick}. Table \ref{tab:stats} shows the dataset statistics.


\begin{table}[!]
\caption{Statistics of Datasets}
\vspace{-5pt}
\label{tab:stats}
\begin{tabular}{c|cccc}
\hline
Dataset                  & \#Nodes                & \#Edges                  & Relation & \#Edges \\ \hline
\multirow{3}{*}{YelpChi} & \multirow{3}{*}{45954} & \multirow{3}{*}{3846979} & R-U-R    & 49315   \\
                         &                        &                          & R-S-R    & 3402743 \\
                         &                        &                          & R-T-R    & 573616  \\ \hline
\multirow{3}{*}{Amazon}  & \multirow{3}{*}{11944} & \multirow{3}{*}{4398392} & U-P-U    & 175608  \\
                         &                        &                          & U-S-U    & 3566479 \\
                         &                        &                          & U-V-U    & 1036737 \\ \hline
\end{tabular}
\vspace{-10pt}
\end{table}

\subsubsection{Baselines}
We are not aware of any similar work addressing SDS. As our focus is GAD, we choose some state-of-the-art methods in the task, some of which utilize the same backbone as us.
\begin{itemize}[leftmargin=*]
    \item \textbf{GCN} \cite{Kipf2017}: GCN is a traditional graph convolutional network.
    \item \textbf{Graph-SAGE} \cite{hamilton2018inductive}: Compared with GCN, Graph-SAGE samples and aggregates features from a node’s local neighborhood, which can generalize to unseen nodes.
    \item \textbf{GAT} \cite{veli2018graph}: A method that leverages masked self-attentional layers to address the shortcomings of prior graph convolution methods.
    \item \textbf{GraphConsis} \cite{liu2020alleviating}: GraphConsis is a heterogeneous graph neural network focusing on tackling context inconsistency, feature inconsistency and relation inconsistency problem.
    \item \textbf{Care-GNN} \cite{dou2020enhancing}: Care-GNN is a camouflage-resistant graph neural network that adaptively samples neighbors according to feature similarity, and the optimal sampling ratio is found through an RL module.
    \item \textbf{FRAUDRE} \cite{zhangfraudre}: This method wants to be dual-resistant to graph inconsistency and imbalance. An additional graph convolution module is leveraged to magnify the difference between normals and anomalies.  
    \item \textbf{PC-GNN} \cite{liu2021pick}: This method consists of two modules ``pick'' and ``choose'', and maintain a balanced label frequency around fraudsters by downsampling and upsampling.
\end{itemize}

\begin{table*}[!]
\caption{Performance Results. Best results of all methods are indicated in bold face, and second best results are underlined.}
\label{tab:res}
\begin{tabular}{c|c|ccc|ccc}
\hline
\multirow{2}{*}{Method}   & Dataset     & \multicolumn{3}{c|}{YelpChi}                                                                                     & \multicolumn{3}{c}{Amazon}                                                                                         \\ \cline{2-8} 
                          & Metric      & \multicolumn{1}{c|}{F1-macro}              & \multicolumn{1}{c|}{AUC}                   & GMean                  & \multicolumn{1}{c|}{F1-macro}               & \multicolumn{1}{c|}{AUC}                    & GMean                  \\ \hline
\multirow{3}{*}{General GNNs} & GCN         & \multicolumn{1}{c|}{0.5171\tiny{±0.0097}}         & \multicolumn{1}{c|}{0.5689\tiny{±0.0157}}         & 0.4541\tiny{±0.0946}          & \multicolumn{1}{c|}{0.6054\tiny{±0.0883}}          & \multicolumn{1}{c|}{0.8667\tiny{±0.0027}}          & 0.7638\tiny{±0.0343}          \\
                          & GraphSAGE   & \multicolumn{1}{c|}{0.4184\tiny{±0.0620}}         & \multicolumn{1}{c|}{0.5400\tiny{±0.0043}}         & 0.3207\tiny{±0.1560}          & \multicolumn{1}{c|}{0.5835\tiny{±0.0088}}          & \multicolumn{1}{c|}{0.7535\tiny{±0.0059}}          & 0.7037\tiny{±0.0076}          \\
                          & GAT         & \multicolumn{1}{c|}{0.5164\tiny{±0.0986}}         & \multicolumn{1}{c|}{0.7403\tiny{±0.0242}}         & 0.6227\tiny{±0.0549}          & \multicolumn{1}{c|}{0.6426\tiny{±0.0359}}          & \multicolumn{1}{c|}{0.8499\tiny{±0.0014}}          & 0.6268\tiny{±0.1265}          \\ \cline{1-8} 
\multirow{4}{*}{GAD models}        & GraphConsis & \multicolumn{1}{c|}{0.6577\tiny{±0.0012}}         & \multicolumn{1}{c|}{0.7853\tiny{±0.0033}}         & 0.6779\tiny{±0.0165}          & \multicolumn{1}{c|}{0.7894\tiny{±0.0448}}          & \multicolumn{1}{c|}{0.9516\tiny{±0.0005}}          & 0.8787\tiny{±0.0025}          \\
                          & FRAUDRE     & \multicolumn{1}{c|}{0.5765\tiny{±0.0482}}         & \multicolumn{1}{c|}{0.7683\tiny{±0.0176}}         & 0.6978\tiny{±0.0195}          & \multicolumn{1}{c|}{0.8682\tiny{±0.0300}}         & \multicolumn{1}{c|}{0.9299\tiny{±0.0035}}          & 0.8768\tiny{±0.0185}          \\
                          & CARE-GNN    & \multicolumn{1}{c|}{0.6433\tiny{±0.0094}}         & \multicolumn{1}{c|}{0.7925\tiny{±0.0292}}         & 0.7094\tiny{±0.0359}          & \multicolumn{1}{c|}{{\ul 0.8988\tiny{±0.0073}}} & \multicolumn{1}{c|}{0.9491\tiny{±0.1115}}          & 0.8908\tiny{±0.0018}          \\
                          & PC-GNN      & \multicolumn{1}{c|}{{\ul 0.6933\tiny{±0.0253}}}   & \multicolumn{1}{c|}{{\ul 0.8512\tiny{±0.0015}}}   & {\ul 0.7720\tiny{±0.014}}     & \multicolumn{1}{c|}{0.8658\tiny{±0.0074}}          & \multicolumn{1}{c|}{{\ul 0.9614\tiny{±0.0014}}}    & {\ul 0.8978\tiny{±0.0044}}    \\ \hline
\multirow{2}{*}{Ours}     & GDN     & \multicolumn{1}{c|}{\textbf{0.7605\tiny{±0.006}}} & \multicolumn{1}{c|}{\textbf{0.9034\tiny{±0.008}}} & \textbf{0.8084\tiny{±0.0009}} & \multicolumn{1}{c|}{\textbf{0.9068\tiny{±0.0042}}}    & \multicolumn{1}{c|}{\textbf{0.9709\tiny{±0.0016}}} & \textbf{0.9078\tiny{±0.0011}} \\ \cline{2-8} 
                          & Improvement & \multicolumn{1}{c|}{9.69\%}                & \multicolumn{1}{c|}{6.13\%}                & 4.72\%                 & \multicolumn{1}{c|}{0.89\%}                & \multicolumn{1}{c|}{0.98\%}                 & 1.11\%                 \\ \hline
\end{tabular}
\end{table*}

\subsubsection{Metrics}
Similar to previous works \cite{liu2021pick}, we adopt three widely used measures for fair comparison: F1-macro, AUC and GMean. F1-macro calculates F1-score for every class and finds their unweighted mean. F1-score is the harmonic mean of precision and recall. AUC is the area under the ROC Curve, which depicts the relationship between False Positive Rate (FPR) and True Positive Rate (TPR). GMean is the geometric mean of True Positive Rate (TPR) and True Negative Rate (TNR). For all of the three metrics, the higher scores indicate the higher performance of the approaches.

\subsubsection{Implementation Details}
The average with standard deviations of 5 runs is reported for all experiments. The backbone of the proposed method is RGCN whose hidden dimension is set to 64. Following previous work, our data splitting ratio is 40\%, 20\%, and 40\% for training, validation, and test set. All of the hyper-parameters are tuned based on the validation set. $\lambda$ is ranged from \{0.01, 0.1, 0.5, 1\}, $k$ for YelpChi is chosen from \{6, 12, 18, 24\}, while that for Amazon is chosen from \{5, 10, 15, 20\}.

\subsection{Comparison Results}

To answer \textbf{RQ1}, we evaluate the performance of baselines and the proposed method, and the comparison results are reported in Table \ref{tab:res}. We implement GCN and Graph-SAGE on our own in Pytorch, and for GraphConsis, CARE-GNN, FRAUDRE, and PC-GNN, we use their provided open-source code to implement them. All of the hyperparameters are set to those reported in their paper if available. We have the following observations:

First of all, CARE-GNN, FRAUDRE, and PC-GNN are three well-designed models for GAD. They are built upon a multi-relation graph, and utilize RGCN as the backbone model as we do. They focus on modifying adjacency matrix to prune noise edges or maintain a balanced neighborhood label frequency, while different from them, we are aiming at learning a powerful and expressive node presentation, which can avoid the high risk of information loss when deleting edges. These three methods are the most competitive baselines so far, and experimental results show that the proposed method consistently outperforms them on all the metrics across two datasets. FRAUDRE performs poorly compared to PC-GNN and CARE-GNN, we think the reason behind this is that FRAUDER deals with the minority class at the end of the topology as a separate module, which may not effectively contribute to the learning process of GNNs. And PC-GNN outperforms CARE-GNN, which is consistent with the conclusion in the paper of PC-GNN.

Secondly, Graph-SAGE is an inductive graph learning method, as it samples a subset of nodes and substantially decreases the difficulty in training on graph for its memory efficiency. However, this sampling technique may also face a high risk for loss of information which leads to performance drop, especially in high imbalanced-heterophily datasets. Therefore, there exists a huge gap between the proposed method GDN and them.

Thirdly, general graph learning methods are evaluated on single-relation graph where all edges are merged. As seen in the table, GAD methods which are built upon multi-relation graph have obvious advantages over GCN, Graph-SAGE and GAT. It demonstrates the effectiveness of treating different relations differently. We think the reason behind is that edges in this manner have a more specific semantic information.  

\subsection{Out-of-Distribution Evaluation}
\begin{figure}[!]
  \centering
  \includegraphics[width=\linewidth]{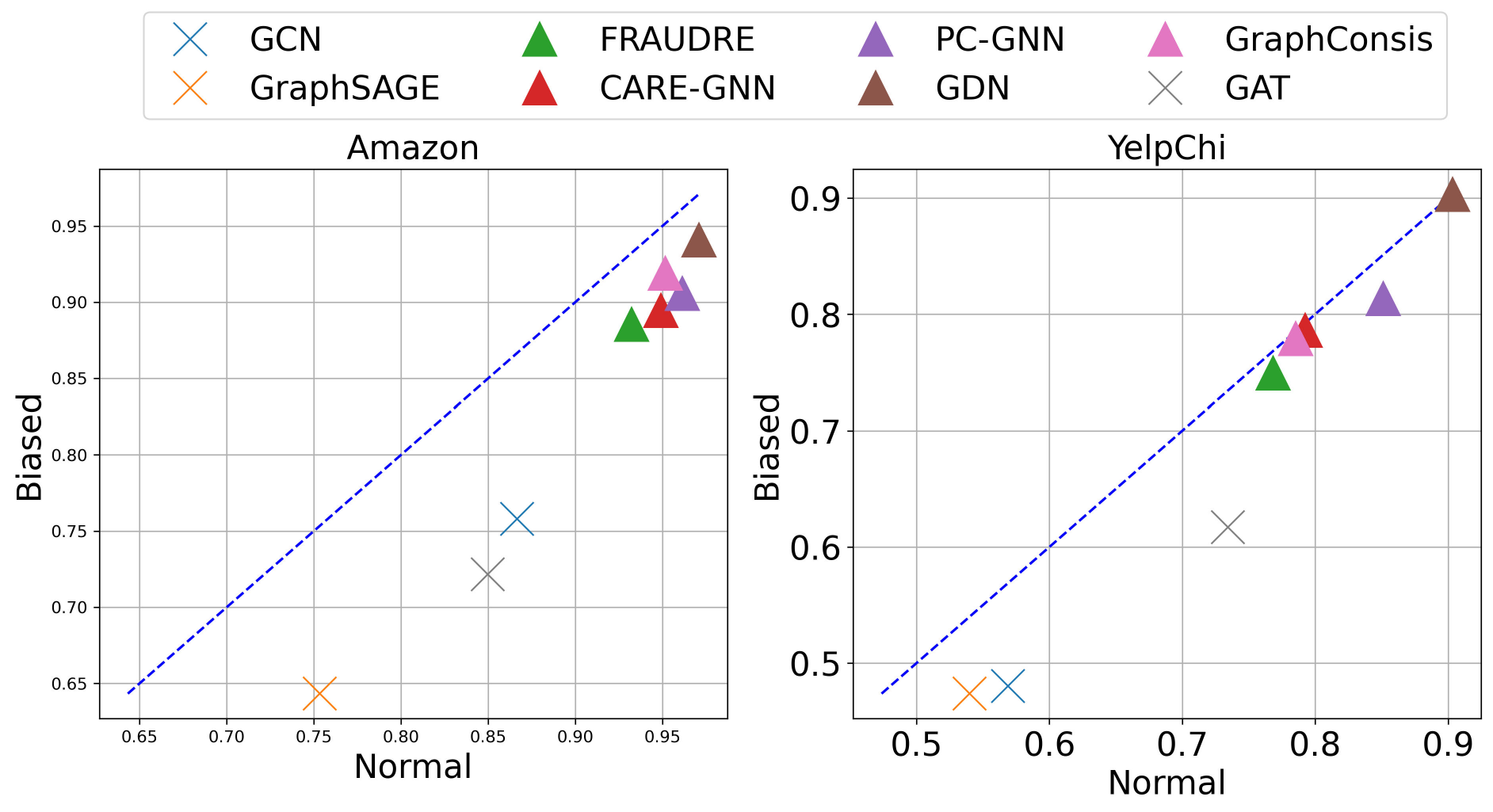}
  \caption{Performance comparison under normal setting and biased setting, the measure is AUC.}
  \label{fig:ood_analysis}
\end{figure}
To answer \textbf{RQ2}, we split datasets according to Equation \eqref{eq:bias_split}. The train-valid-test split ratio remains the same with the normal setting, and note that we treat both train nodes and validation nodes as \textit{observed}, \ie their neighborhood distributions are the same. We suppose this processing could simulate the real-world better, and help with fair comparison when doing hyper-parameter tuning. We conduct experiments on the proposed method and baselines. From the result reported in Figure \ref{fig:ood_analysis}, we can observe that: 

The proposed method outperforms all of the baselines under both biased and normal settings, which again demonstrates the effectiveness of our model. In addition, multi-relation methods marked by triangles perform much better than single-relation methods marked by ``X''. Next, as compared between two figures in \ref{fig:ood_analysis}, it is obvious that SDS (indicated by the distance from points to the dashed blue line $y=x$) is more severe on Amazon than that on YelpChi, we attribute this difference to the difference in heterophily of anomalies. From Figure \ref{visual_SDS}, we can see the heterophily degree is higher on Amazon than that on YelpChi. This finding is consistent with our assumption that SDS is exaggerated by heterophily. Since GDN's performance gap between biased setting and normal setting on both benchmarks is small compared with other GAD methods, our model has the promising ability to alleviate SDS. 

\begin{table}[!]
\caption{LR and LP prediction results with different feature combinations. The measure is AUC.}
\vspace{-5pt}
\label{tab:fs}
\begin{tabular}{cc|c|c|c|c}
\hline
\multicolumn{2}{c|}{}                             & $C$      & $C^{'}$     & $C+S$    & $(C+S)^{'}$ \\ \hline
\multicolumn{1}{c|}{\multirow{2}{*}{LR}} & Amazon & 0.9069 & 0.9047 & 0.9084 & 0.9062 \\ \cline{2-6} 
\multicolumn{1}{c|}{}                    & YelpChi   & 0.7314 & 0.7432 & 0.7545 & 0.7694 \\ \hline
\multicolumn{1}{c|}{\multirow{2}{*}{LP}} & Amazon & 0.9277 & 0.9289 & 0.9293 & 0.9298 \\ \cline{2-6} 
\multicolumn{1}{c|}{}                    & YelpChi   & 0.7285 & 0.7377 & 0.7418 & 0.7447 \\ \hline
\end{tabular}
\vspace{-10pt}
\end{table}
\subsection{Feature Separation Analysis}

\begin{figure}[!]
     \centering
     \begin{subfigure}[b]{0.5\linewidth}
         \centering
         \includegraphics[width=\linewidth]{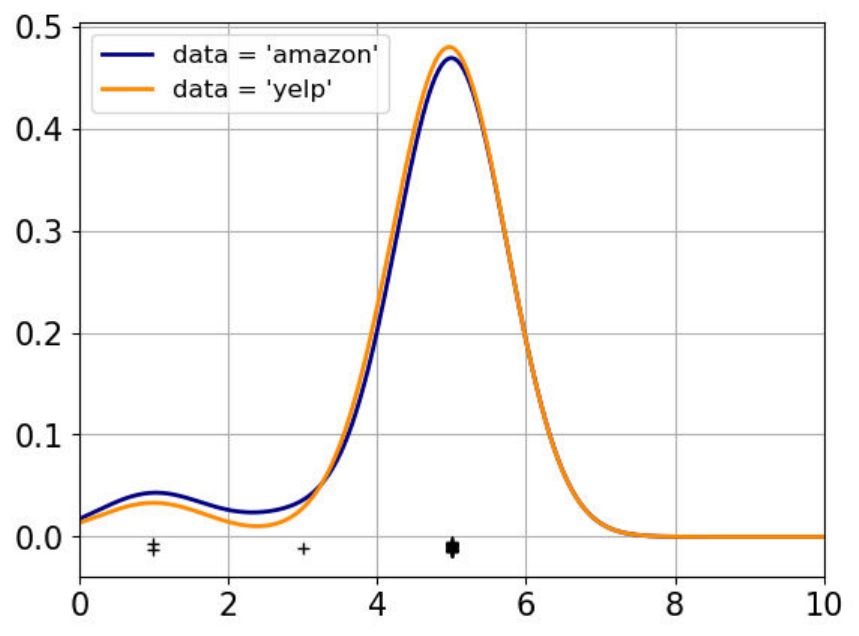}
         \caption{}
         \label{fig:2d_grad}
     \end{subfigure}%
     \begin{subfigure}[b]{0.5\linewidth}
         \centering
         \includegraphics[width=\linewidth]{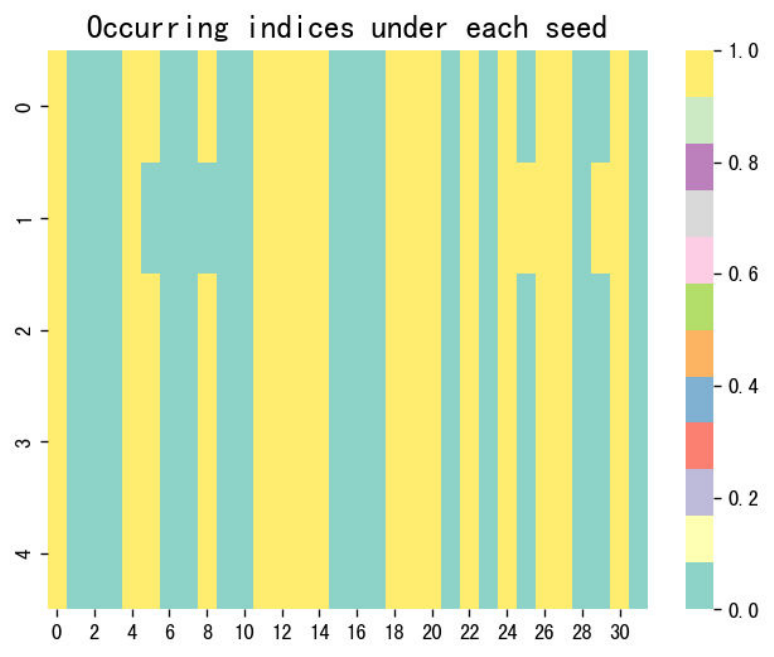}
         \caption{}
         \label{fig:3d_grad}
     \end{subfigure}
     \caption{Feature Separation Stability Analysis.}
     \label{fig:gradient_analysis}
\vspace{-15pt}
\end{figure}

The key to the biased classification performance is feature separation. To take a closer look in \textbf{RQ2}, we verify the effectiveness of our feature separation module from both the data perspective and model perspective. Inspired by \cite{9811416}, we first visualize the frequency distributions of chosen class features $C$ on 5 different random seeds. Results are reported in Figure \ref{fig:gradient_analysis}. In Figure \ref{fig:gradient_analysis}(a), we adopt kernel density estimation (KDE) and fit the shape of distribution with Gaussian Kernel; in Figure \ref{fig:gradient_analysis}(b), a detailed index constitution of class feature for each seed is displayed. From the figure, we observe the framework always selects the same feature dimension, and our separation process is stable and constant. From model perspective, we aim to verify the assumption that $C$ inherits most of the samples' informative characteristics, and $S$ improves the performance of graph-based methods. Towards this end, we measure the performance of different feature combinations on simple linear classifiers and graph-based classifiers, because we suppose that the simpler the classifier, the better it reflects the feature quality. We train Logistic Regression classifiers and Label Propagation classifiers on original class features ($C$), regularized class features ($C^{'}$), original features ($C+S$), and regularized features ($(C+S)^{'}$), respectively. Note that LP algorithm reconstructs the graph according to nodes' similarity and is trained on the new graph instead of the original graph. Therefore, LP is less likely to suffer from class heterophily, whose performance is a good measurement of nodes' ability to represent their local structure. Results are displayed in Table \ref{tab:fs}. From the table, we have the following observations:

Comparing Table \ref{tab:fs} with Table \ref{tab:res}, the linear model (LR) achieves better performance than general GNN-based methods like GAT, GraphSAGE and GCN. It is consistent with previous work and our assumption that graph-based methods suffer from imbalanced heterophily problem in GAD. Secondly, $C$ has little performance drop on LR compared with $C+S$, suggesting that $C$ has the ability to represent most information contained in the whole feature set. What's more, $C+S$ outperforms $C$ on LP, indicating that including $S$ in the feature set is conducive to the performance of graph-based methods. In addition, $C^{'}$ and $(C+S)^{'}$ achieves better or comparable performance to $C$ and $C+S$, which demonstrates the effectiveness of our regularization terms.  


\subsection{Hyper-parameter Sensitivity Analysis}
\begin{figure*}[h]
     \centering
     \begin{subfigure}[b]{0.25\textwidth}
         \centering
         \includegraphics[width=\textwidth]{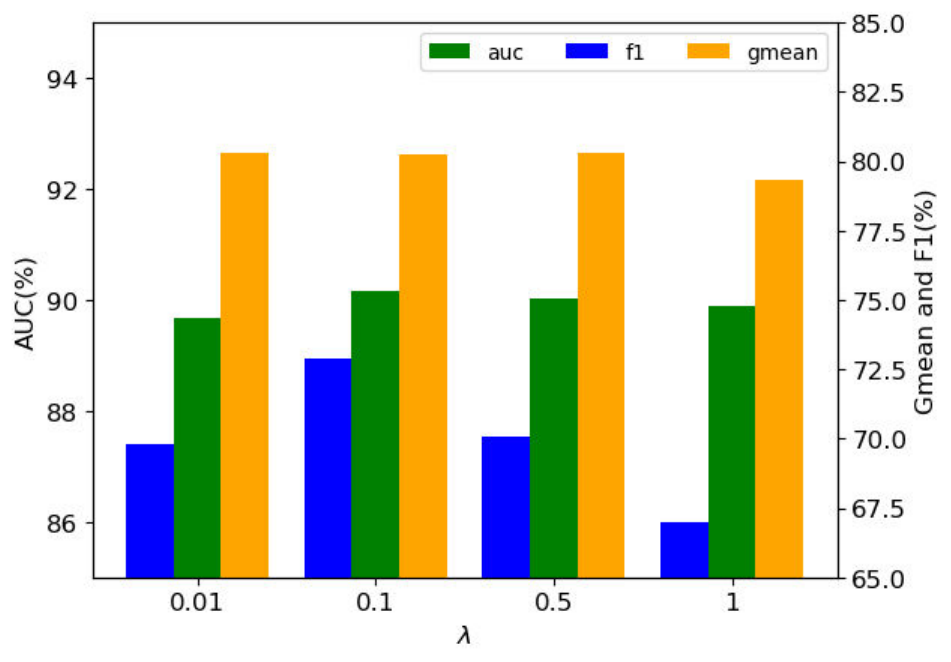}
         \caption{YelpChi \wrt $\lambda$}
         \label{fig:yelp_lambda}
     \end{subfigure}%
    \begin{subfigure}[b]{0.25\textwidth}
         \centering
         \includegraphics[width=\textwidth]{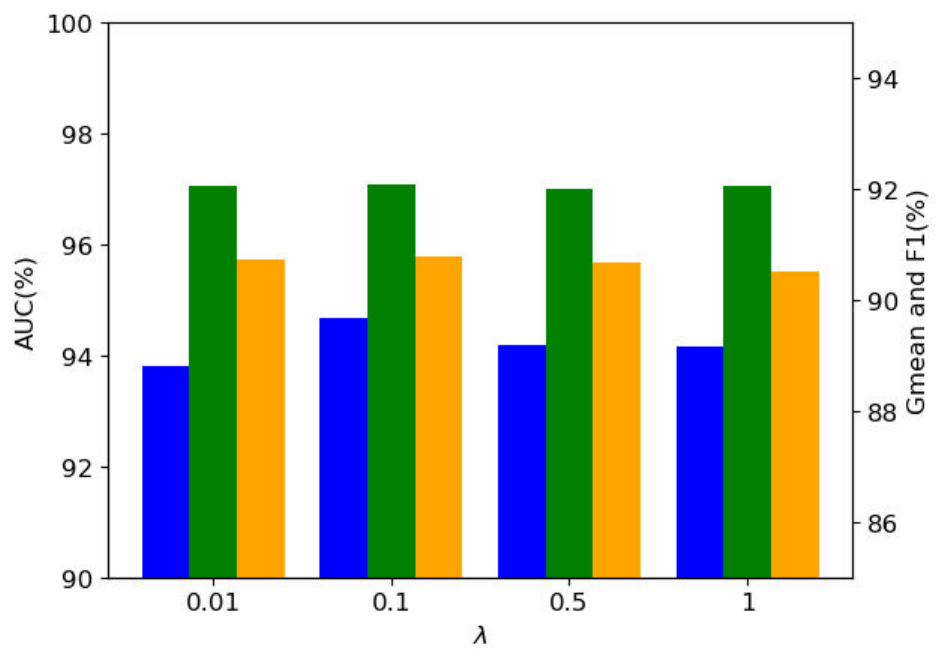}
         \caption{Amazon \wrt $\lambda$}
         \label{fig:amazon_lambda}
     \end{subfigure}%
     \begin{subfigure}[b]{0.25\textwidth}
         \centering
         \includegraphics[width=\textwidth]{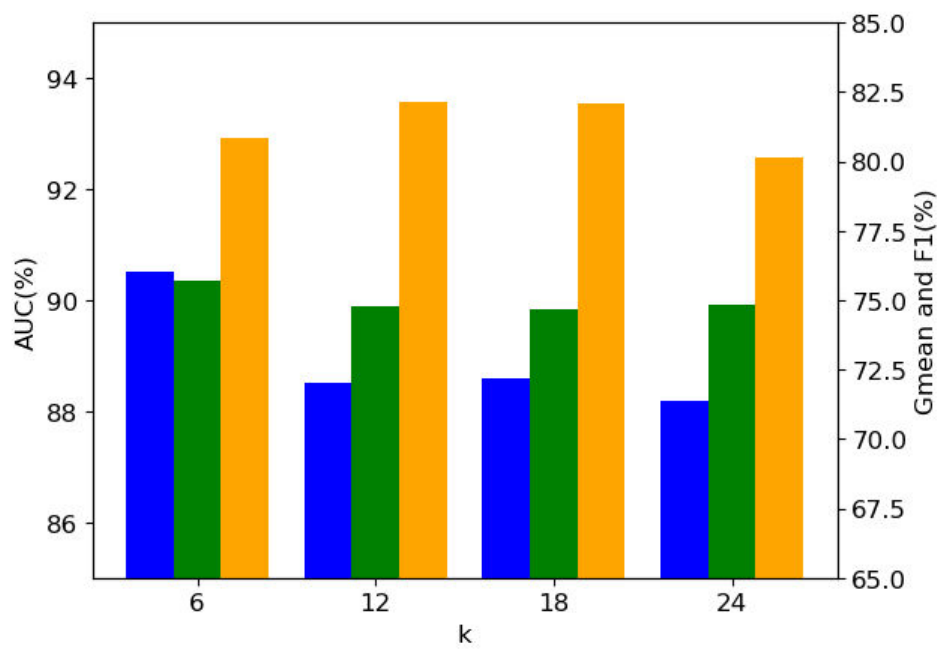}
         \caption{YelpChi \wrt k}
         \label{fig:yelp_k}
     \end{subfigure}%
     \begin{subfigure}[b]{0.25\textwidth}
         \centering
         \includegraphics[width=\textwidth]{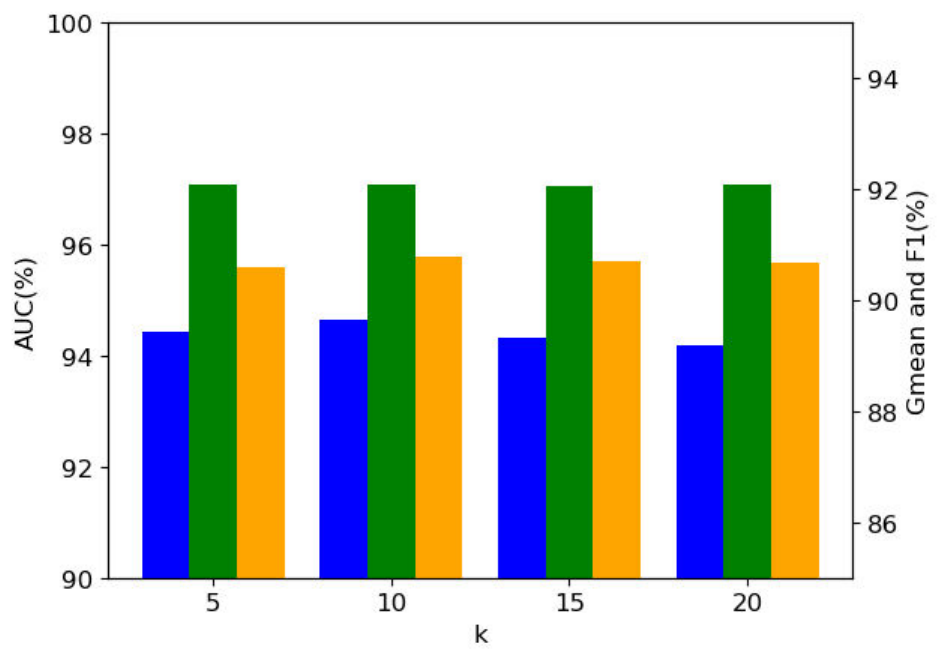}
         \caption{Amazon \wrt k}
         \label{fig:amazon_k}
     \end{subfigure}
     \caption{Model performance with different hyper-pamameters on YelpChi and Amazon Dataset. Since three metrics are different in scales, we display the charts in a dual-Y style.}
     \label{fig:sensitivity}
\end{figure*}

To answer \textbf{RQ3}, we want to explore the model's sensitivity to the most important parameters $\lambda$ and $k$, which control the integration of two losses and the dimensions of $C$ and $S$ when conducting feature separation. The results of three metrics with different $\lambda$ and $k$ are shown in Figure \ref{fig:sensitivity}, respectively. We observe that (1) Generally, when $\lambda$ continues to increase, the performance will first increase then decrease. We suppose a small $\lambda$ has an inadequate influence on the classification while the classifier may be dominant by the auxiliary regularization when it is too large. (2) Similarly, with the increase of $k$, the performance tends to first increase then decrease. (3) The performance is stable over a quite large range.

\subsection{Flexibility Analysis}
To further verify the flexibility of GDN and answer \textbf{RQ4}, we leverage feature separation module and regularization terms on several representative GNNs and observe large improvement on both of the datasets. Experiment results are reported in Table \ref{tab:flexibility}. From table \ref{tab:flexibility}, we can conclude that our proposed method can be easily adapted to general models and enhance their performance.

 \section{Related Work}
 In this section, we introduce some representative previous works on graph-based anomaly detection and out-of-distribution learning.

 \subsection{Graph Anomaly Detection}
 In GAD, researchers tend to categorize the problem into two branches based on whether the graph data is time-invariant \cite{akoglu2015graph, ma2021comprehensive}. We focus on summarizing existing GAD works on static graphs.
 
 On static attributed graphs, auto-encoder (AE) based methods can not handle the graph data directly. DONE \cite{bandyopadhyay2020outlier} trains two AEs by minimizing the reconstruction error while preserving the homophily between connected nodes. With the advances of GNNs, GNN-based methods \cite{ding2019deep, zhang2020gcn, liang2018semi} have been of focus. GraphRfi \cite{zhang2020gcn} explores the possibility of combining anomaly detection with downstream graph tasks. GraphUCB \cite{ding2019interactive} adopts contextual multi-armed bandit technology, and transform graph anomaly detection to a decision-making problem. DCI \cite{wang2021decoupling} decouples representation learning and classification with the self-supervised learning task.
 
 
 Recent methods realize the importance of incorporating multiple relationships into graph learning \cite{wang2019semi, wang2019fdgars, liu2020alleviating, dou2020enhancing, liu2021intention, liu2021pick}. FdGars \cite{wang2019fdgars} and GraphConsis \cite{liu2020alleviating} construct a homo-graph with multiple relations and leverage GNNs to aggregate neighborhood information. Differently, Semi-GNN \cite{wang2019semi}, CARE-GNN \cite{dou2020enhancing}, and PC-GNN \cite{liu2021pick} construct multiple homo-graphs based on node relation. Semi-GNN and IHGAT \cite{liu2021intention} employ hierarchical attention mechanism for interpretable prediction, while CARE-GNN and PC-GNN prune edges adaptively according to neighbor distribution.
 
 \begin{table}[!]
\caption{Enhancement for other models. The measure is AUC.}
\label{tab:flexibility}
\begin{tabular}{c|c|c|c|c}
\hline
       & GCN    & GCN+GDN & SAGE   & SAGE+GDN \\ \hline
Amazon & 0.8667 & 0.8904    & 0.7535 & 0.8216     \\ \hline
YelpChi   & 0.5689 & 0.5832    & 0.5400 & 0.7467     \\ \hline
\end{tabular}
\end{table}

 \subsection{Learning under Out-of-Distribution}
 In OOD problem, the classic methods cannot be directly applied. Existing algorithms for OOD generalization can be divided into unsupervised learning and supervised learning \cite{shen2021towards}. 
 
 Unsupervised representation learning methods are mainly based on VAE. $\beta$-VAE \cite{higgins2016beta} introduces an learnable hyperparameter $\beta$, balancing latent channel capacity and independence constraints. FactorVAE \cite{kim2018disentangling} improves $\beta$-VAE by disentangling which encourages the representation distribution to be factorial and independent. CausalVAE \cite{yang2021causalvae} leverages a causal layer that transforms independent exogenous factors into causal endogenous ones. DEAR \cite{shen2020disentangled} uses a structural causal model (SCM) as the prior for a bidirectional generative model, and a suitable GAN loss is introduced as supervision to train a generator and an encoder jointly.
 
 Supervised model learning wants to learn a robust model that can be generalized to the unseen target domain. CIAN \cite{li2018deep} leverages DNN to learn invariance representation with respect to the joint distribution of representation learning function and feature. Graph-DVD \cite{shen2020stable} decorrelates the stable feature for classification from unstable variables. Some works \cite{peters2016causal, chow1960tests, pfister2019invariant, heinze2018invariant} based on ICP \cite{peters2016causal} want to connect invariance to causality, which performs a statistical test on whether the invariance assumption is met.
 
 However, the OOD problem is under-explored in graph. SR-GNN \cite{zhu2021shift} and an IRM-based method \cite{wu2022towards} notice OOD problems in node feature distribution, and a PAC-Bayesian analysis \cite{ma2021subgroup} demonstrates non-IID data can affect the performance of subgroups. In this work, we suppose that neighborhood information also suffers from SDS. 
 
\section{Conclusion and Future Work}

In this work, we explain a novel problem --- structural distribution shift in GAD. To alleviate SDS, we propose a novel method GDN. GDN separates node features into two parts: one ascertain an invariant pattern for anomalies, while the other endows with the ability to benefit from aggregation mechanism for normals. 

The method takes the first step to address structural distribution shift in GAD. For future work, there are some research directions worth studying: 1) Integration of feature constraint and edge pruning. Noisy edges are truly harmful, an OOD-guided edge pruning method deserves our attention. Also a mechanism to integrate both algorithms is helpful. 2) Better regularizer. The discovery of a better distance function for regularization terms, such as $f$-divergence and Wasserstein distance, since $KL$-divergence has some limitations. 3) Spectral domain. Some GAD works \cite{DBLP:conf/icml/TangLGL22,yangspectral} recently published find that anomaly nodes can incur high frequency, addressing SDS in spectral domain is another future direction. 

\section{Acknowledgments}
This work is supported by the National Key Research and Development Program of China (2021ZD0111802), the National Natural Science Foundation of China (62121002, U1936210, 9227010114), and the CCCD Key Lab of Ministry of Culture and Tourism.

\bibliographystyle{ACM-Reference-Format}
\bibliography{sample-base}
\end{document}